\documentclass[letterpaper, 10 pt, conference]{ieeeconf}  

\IEEEoverridecommandlockouts                              

\usepackage{booktabs}
\usepackage{graphicx}
\usepackage[table]{xcolor}
\usepackage{amsmath,amssymb}
\usepackage{tabularx}
\usepackage{array}
\usepackage{xurl}
\usepackage{placeins}
\definecolor{paperlink}{HTML}{433899}

\makeatletter
\let\NAT@parse\undefined
\makeatother

\usepackage[
    colorlinks=true,
    linkcolor=paperlink,
    citecolor=paperlink,
    urlcolor=paperlink
]{hyperref}
\title{\LARGE \bfseries
iGPC: Generative Motion Priors for Object-Aware \\ Humanoid Interaction
}

\author{%
Anujith Muraleedharan$^{1}$, Abdul Ahad Butt$^{1}$, Nolan Fey$^{2}$, Yash Prabhu$^{2}$, \\ Anamika J H$^{1}$, Sandor Felber$^{2}$, Maurice Rahme$^{2}$, Ivan Laptev$^{1}$%
\thanks{$^{1}$Mohamed bin Zayed University of Artificial Intelligence (MBZUAI),
Abu Dhabi, United Arab Emirates.}%
\thanks{$^{2}$Minerva Humanoids, Inc.}%
}

\begin{document}

\maketitle
\thispagestyle{empty}
\pagestyle{empty}

\begin{abstract}

Humanoid robots operating in unstructured environments must combine robust whole-body control with the ability to perceive and physically interact with surrounding objects. While large-scale human motion data provides powerful priors for natural and versatile humanoid control, effectively transferring such priors to perception-driven object interaction remains challenging. To address this bottleneck, we propose a framework that extends the recently proposed Generative Pretrained Controller (GPC) from general human motion to full-body humanoid–environment interaction. First, we adapt GPC into interaction experts conditioned on scene affordance cues and privileged state information. This adaptation begins with interaction
and affordance pre-training on human motions paired with support
scenes. These experts leverage the pretrained human motion prior while learning task-specific contact behaviors, including reaching toward objects, grasping environmental supports for stabilization, and pushing movable objects. Second, we introduce a perception-driven student that retains the pretrained GPC policy and distills interaction skills from the experts using onboard sensory observations. 
To bridge the gap between privileged expert
observations and sensory inputs, we train the student's perception
not only to reconstruct expert observations, but also to preserve
the expert's motion decisions and resulting joint targets. Notably,
our experiments across multiple whole-body interaction tasks
demonstrate that large-scale generative human motion priors provide
an effective foundation for learning deployable policies for humanoid
interactions in contact-rich real-world environments.

\end{abstract}

\section{Introduction}
\label{sec:introduction}

Recent advances in humanoid control have demonstrated that
large-scale human motion data can provide powerful priors
for learning natural and versatile whole-body behaviors.
Motion-tracking approaches such as ZEST and BeyondMimic
transfer diverse human motions to humanoid
robots~\cite{sleiman2026zest,liao2025beyondmimic}, while
generalist controllers such as GMT and SONIC scale this
paradigm to increasingly diverse motion repertoires and
command modalities~\cite{chen2025gmt,luo2025sonic}.
More recently, generative approaches have gone beyond motion
replay by learning reusable distributions over motor behaviors.
In particular, GPC~\cite{shi2026gpc} learns an autoregressive
generative motor prior from large-scale motion data and
adapts this prior to downstream control tasks.
These results suggest an appealing alternative to learning
individual humanoid skills from scratch: pretrain a general
motor prior once and adapt it to increasingly complex
downstream behaviors.

\begin{figure}[!t]
    \centering
    \includegraphics[width=\columnwidth]{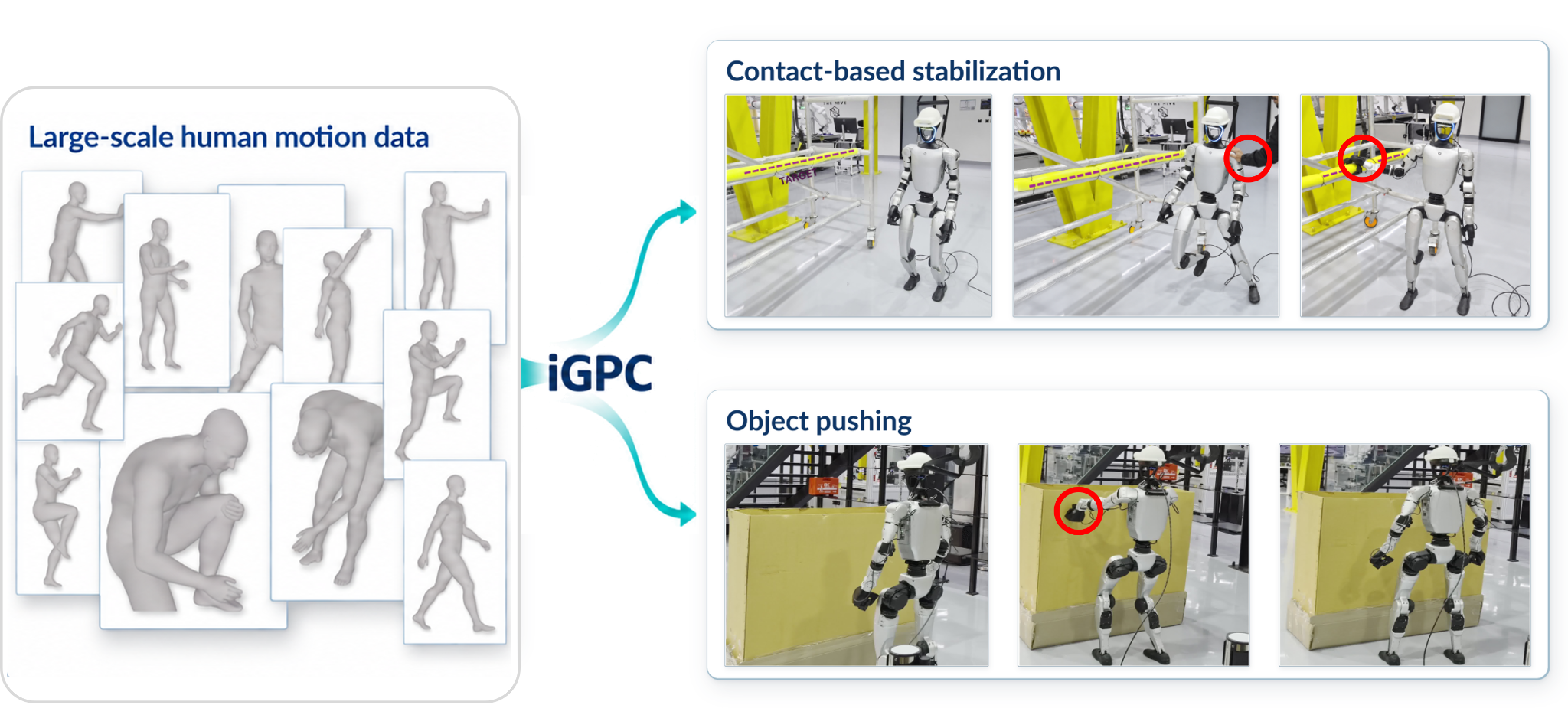}
    \caption{
        iGPC adapts human-motion priors to perceptive
        whole-body interaction on Unitree G1.
        Real-world sequences illustrate rail-assisted stabilization
        following an external push (top) and box pushing
        (bottom).
    }
    \label{fig:teaser_igpc}
\end{figure}
A key challenge, however, is extending such motion priors from free-space whole-body control to physical interaction with the environment. Reaching for an object, grasping a support to maintain balance, or pushing a heavy object requires the robot to coordinate locomotion, posture, and contact according to its surroundings. Human motion priors provide useful knowledge about coordinated whole-body movement, but do not by themselves tell the robot {\em where and how to interact with a particular scene}. 

Recent approaches address perceptive humanoid control through terrain-conditioned motion tracking~\cite{wang2026perceptivebfm,li2026pgmt}, reference-conditioned object interaction~\cite{zhao2025resmimic,chen2026scenebot}, or perception-based loco-manipulation~\cite{he2026ultra,lin2026lessmimic,krishna2026vibe}. Nevertheless, it remains unclear how a generative motor prior learned from large-scale human motion can be systematically adapted to object-aware interaction and subsequently retained when the controller must operate from onboard sensing rather than privileged state information.

We address this problem with {\em iGPC}, a framework that extends {\em GPC} from generative motion control to object-aware, perception-driven humanoid {\em interaction}. First, we develop an improved interaction expert that conditions the GPC motion prior on scene affordances describing object geometry, contact, and robot stability. We initialize adaptation using human interaction motions aligned with scene geometry and subsequently optimize task-specific behaviors with reinforcement learning. The resulting experts exploit large-scale human motion priors while learning task-specific interactions, including reaching, grasping environmental supports for stabilization, and pushing movable objects, without requiring reference motions at execution time.

To enable real-world deployment, we propose a perceptive student
that retains the same GPC motion prior rather than replacing it
with a separately learned action policy. The student uses onboard
depth, LiDAR, and proprioception to estimate the interaction and
body-state context required by GPC. We introduce distillation
objectives that encourage the student to reproduce both the
expert's motion decisions, represented by motor tokens, and the
resulting joint targets. To reduce the effect of imperfect sensor
estimates on these decisions, we learn to adjust the estimated
information passed to the motion prior. We then further optimize
the student's task behavior through reinforcement learning.
Consequently, the large-scale human motion prior benefits both
the experts and perceptive controllers.

Experiments demonstrate strong improvements at both the expert and
student levels. For object-aware reaching, the iGPC expert achieves
99.32\% success in contacting the upper surface of a rail with the
commanded hand, compared with 31.30\% for GPC, while the sensor-based
student retains 95.95\% success and substantially reduces body
collisions relative to the expert. The approach also improves hand-supported recovery from strong disturbances compared with the GPC baseline. For sustained box pushing, both the expert and student
outperform VisualMimic~\cite{yin2025visualmimic}, achieving approximately
46--50\% greater mean forward displacement and 62--65\% lower mean
absolute lateral drift.
Together, these results show that generative human motion priors
provide an effective foundation for both privileged interaction
learning and perception-driven humanoid control.

In summary, our main contributions are threefold:

\begin{itemize}
    \item {\bf Object-aware interaction experts.} We extend GPC with scene affordance, contact, and stability conditioning, enabling motion-prior-based reaching, support-assisted stabilization, and object manipulation without reference motions at execution time.
    
    \item {\bf Motion-prior-based perceptive control.} We introduce a multimodal distillation framework that transfers privileged experts to onboard depth, LiDAR, and proprioception while retaining the pretrained GPC motor prior and decoder.

    \item {\bf Extensive experimental validation.} We demonstrate strong performance of iGPC across reaching, stabilization, and box pushing, with ablations validating our expert and student designs and comparisons showing substantial improvements over the baselines.

\end{itemize}

\section{Related Work}
\label{sec:related_work}

\noindent
{\bf Reusable motor representations.}
Motion priors provide coordinated behavior that can be reused for
new objectives. BeyondMimic combines tracking with guided diffusion
for goal-directed motion composition, while SONIC connects
large-scale tracking to planning and multiple command
modalities~\cite{liao2025beyondmimic,luo2025sonic}. BFM-Zero supports
task prompting through a behavioral latent representation, and
LATENT learns high-level tennis control over a motion-derived
latent action space~\cite{li2025bfmzero,zhang2026latent}. GPC instead
models discrete motor skills autoregressively and introduces
conditional low-rank adaptation with supervised and reinforcement
learning for physics-based character
control~\cite{shi2026gpc}. We adopt this foundation-training formulation for G1 because it supports motion-reference-free task execution and low-rank
adaptation while retaining the motor backbone and decoder.
This modular structure lets us transfer interaction experts to
onboard sensing by learning their input interface rather than
replacing the motor policy. Our focus is extending the resulting motor
representation to object-aware task control using sensor inputs.

\noindent
{\bf Object-aware and perceptive whole-body control.}
Perceptive BFM and PGMT address terrain-compatible motion tracking
by incorporating local geometry~\cite{wang2026perceptivebfm,li2026pgmt}.
SceneBot reconstructs scenes around retargeted motions and supplies
per-link contact commands~\cite{chen2026scenebot}; ResMimic learns
action residuals around a general tracker to follow coupled
humanoid--object trajectories~\cite{zhao2025resmimic}. These
formulations address interaction while retaining motion-reference
inputs. ViBe also retains a reference-conditioned tracker, learning
visual extraction and low-rank decoder adaptations around frozen
visual and motor backbones~\cite{krishna2026vibe}. SplitAdapter
targets a different source of variation: object load and robot
dynamics, represented by factorized context encoders that modulate
a frozen manipulation policy~\cite{kang2026splitadapter}.
For reactive support, Bracing for Impact uses LiDAR-informed wall
bracing for push recovery and is evaluated in
simulation~\cite{yang2025bracing}.
Our task-specific adapters condition a shared generative motor
foundation on commands, body-relative surface geometry, contact,
and stability, without requiring a prescribed motion reference
at execution. 
\begin{table}[!t]
    \centering
    \caption{Qualitative comparison of selected methods.}
    \label{tab:related_work_comparison}

    \begingroup
    \footnotesize
    \setlength{\tabcolsep}{0.65pt}
    \renewcommand{\arraystretch}{1.20}
    \renewcommand{\tabularxcolumn}[1]{m{#1}}

    \begin{tabularx}{\columnwidth}{
        @{}
        >{\raggedright\arraybackslash}X
        >{\centering\arraybackslash}m{0.170\columnwidth}
        >{\centering\arraybackslash}m{0.155\columnwidth}
        >{\centering\arraybackslash}m{0.140\columnwidth}
        >{\centering\arraybackslash}m{0.130\columnwidth}
        >{\centering\arraybackslash}m{0.140\columnwidth}
        @{}
    }
        \toprule

        \textbf{Method}
        &
        {\scriptsize\bfseries
         \shortstack{Object\\manipulation}}
        &
        {\scriptsize\bfseries
         \shortstack{Reactive\\hand-\\supported\\stabilization}}
        &
        {\scriptsize\bfseries
         \shortstack{Onboard-\\perceptive\\contact}}
        &
        {\scriptsize\bfseries
         \shortstack{Motion-\\reference-\\free\\execution}}
        &
        {\scriptsize\bfseries
         \shortstack{Frozen-\\backbone\\low-rank\\adaptation}}
        \\

        \midrule

        GPC \cite{shi2026gpc}
        & --
        & --
        & --
        & $\checkmark$
        & $\checkmark$
        \\

        BeyondMimic \cite{liao2025beyondmimic}
        & --
        & --
        & --
        & $\checkmark$
        & --
        \\

        ResMimic \cite{zhao2025resmimic}
        & $\checkmark$
        & --
        & --
        & --
        & --
        \\

        ULTRA \cite{he2026ultra}
        & $\checkmark$
        & --
        & $\checkmark$
        & $\checkmark$
        & --
        \\

        LessMimic \cite{lin2026lessmimic}
        & $\checkmark$
        & --
        & $\checkmark$
        & $\checkmark$
        & --
        \\

        ViBe \cite{krishna2026vibe}
        & $\checkmark$
        & --
        & $\checkmark$
        & $\circ^{\mathrm{v}}$
        & $\checkmark$
        \\

        Bracing for Impact \cite{yang2025bracing}
        & --
        & $\checkmark^{\mathrm{s}}$
        & $\checkmark^{\mathrm{s}}$
        & $\checkmark$
        & N/A
        \\

        \midrule

        \textbf{Ours}
        & $\checkmark$
        & $\checkmark$
        & $\checkmark$
        & $\checkmark$
        & $\checkmark$
        \\

        \bottomrule
    \end{tabularx}

    \vspace{3pt}

    \begin{minipage}{\columnwidth}
    \footnotesize
    \raggedright
    $\checkmark$: reported; --: not reported.
    $^{\mathrm{s}}$Simulation only.
    $\circ^{\mathrm{v}}$Partial: ViBe uses a nominal standing
    reference for dodgeball; locomotion and manipulation
    remain reference-conditioned.
\end{minipage}

    \endgroup
\end{table}

\noindent
{\bf Privileged-to-sensor transfer.}
ULTRA distills a tracker into a multimodal variational controller
using action matching, latent regularization, and reconstruction
for tracking and sparse-goal execution~\cite{he2026ultra}.
LessMimic combines reference-free interaction with depth-based
DAgger distillation, replacing explicit geometric inputs with
visual features while retaining the control-head
architecture~\cite{lin2026lessmimic}. VIGOR supervises both recovery
representations and actions for perceptive fall
safety~\cite{azulay2026vigor}. Retaining motor components while
learning their inputs is also established: RMA estimates an
environment latent for a retained base
policy~\cite{kumar2021rma}, and AnyBody adapts a sparse-keypoint
encoder through latent and action supervision with a frozen
decoder~\cite{li2026anybody}. Generalized on-policy distillation supervises learner-generated
sequences~\cite{agarwal2024gkd}; we apply this principle to motor
tokens within each control step. Our transfer method reconstructs
expert observations and supervises motor-token decisions and joint
targets through a retained autoregressive controller. It separately
corrects the inputs used for token selection while preserving the
physical-state estimate supplied to the decoder.
Table~\ref{tab:related_work_comparison} compares selected methods
by object manipulation, reactive hand support, onboard perception,
motion-reference dependence, and motor-backbone adaptation.

\begin{figure*}[!t]
    \centering
    \includegraphics[width=\textwidth]{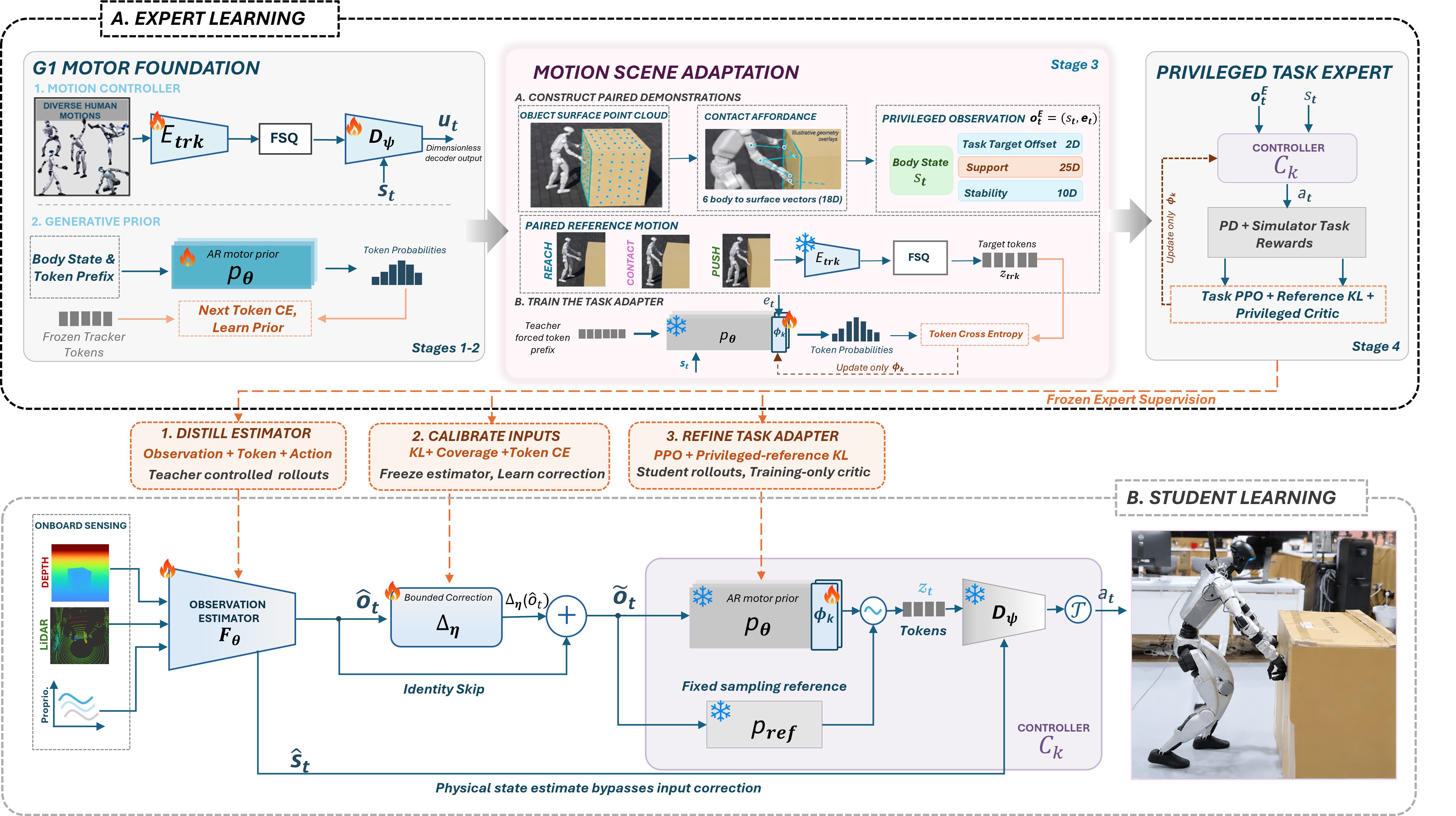}
    \caption{
        Overview of the proposed generative control framework.
        \textbf{(A) Expert learning.}
        A G1-specific motion controller and autoregressive motor
        prior form the motor foundation. Motion--scene supervision
        initializes object-conditioned task adapters, which are
        subsequently refined using privileged simulation observations. 
        \textbf{(B) Student learning.}
        Depth, LiDAR, and proprioceptive history reconstruct the
        expert's observation interface. Observation, token, and
        action supervision train the estimator, followed by bounded
        input calibration and task-adapter refinement.
        Calibrated observations $\tilde{o}_t$ condition token
        selection, while the physical-state estimate $\hat{s}_t$
        bypasses this correction and directly conditions the
        decoder $D_\psi$. Flames indicate modules trained in the corresponding
step; snowflakes indicate frozen learned modules. Dashed orange arrows indicate
        training supervision.
    }
    \label{fig:method_overview}
\end{figure*}

\section{Problem Formulation}
\label{sec:problem_formulation}

\noindent

\paragraph{\textbf{Task-conditioned whole-body interaction}}
We model task-conditioned whole-body interaction on a Unitree G1
as an episodic partially observable Markov decision process.
The hidden state $x_t$ comprises robot configuration and
velocities, object states, and contact conditions. Dynamics
$x_{t+1}\sim P(\cdot\mid x_t,a_t)$ include contact interactions
and external disturbances. The learned action
$a_t\in\mathbb{R}^{29}$ specifies joint-position targets for
low-level proportional--derivative control; Dex3 finger control
is separate (Sec.~\ref{sec:experimental_setup}). Task conditions
are implicit in the policy and reward notation. For each task
$k$, we seek a policy $\pi_k$ that maximizes

\begin{equation}
    J_k(\pi_k)
    = \mathbb{E}_{\pi_k,P}\!\left[
        \sum_{t=0}^{T-1}\gamma^t r_k(x_t,a_t)
    \right],
    \label{eq:interaction_objective}
\end{equation}
where $T$ is the episode horizon and $\gamma\in[0,1)$ is the
discount factor. Task-specific rewards encourage task progress,
useful contact, and stable execution while penalizing undesirable
physical interaction.

\noindent
\paragraph{\textbf{Training and deployment observations}}
During simulation training, the expert receives
$o_t^E=(s_t,e_t)$, where $s_t$ denotes the body-state
representation and $e_t$ combines task-specific inputs
with the support descriptor $g_t$ and stability
descriptor $b_t$.
The deployed student acts according to
\begin{equation}
    a_t\sim\pi_k(\cdot\mid\mathcal I_t),
    \label{eq:sensor_policy}
\end{equation}
where $\mathcal I_t$ includes current depth and LiDAR
measurements, finite proprioceptive and previous-action
history, onboard state estimates, sensor status, and
supplied task conditions. Recurrent encoders process
the finite history window without carrying persistent
hidden state between policy calls. Privileged object
poses, simulator body velocities, and contact-force labels
are not inputs to the learned student body policy.


\section{Method}
\label{sec:method}

iGPC combines object-aware expert learning
(Sec.~\ref{sec:object_conditioned_control}) with transfer to
onboard sensing (Sec.~\ref{sec:sensor_transfer}), retaining
the GPC motor backbone and decoder
(Fig.~\ref{fig:method_overview}). The student reconstructs
expert inputs and matches motor-token decisions and joint
targets, followed by prior-input calibration and task-adapter
refinement.

\subsection{Object-Aware Interaction Experts}
\label{sec:object_conditioned_control}

\noindent {\bf Retained motor foundation.}
We use a G1-specific motor foundation trained following
GPC~\cite{shi2026gpc}, retaining its finite scalar
quantization (FSQ) representation ~\cite{mentzer2023fsq}, autoregressive prior,
and conditional low-rank adaptation mechanism.
Let $E_{\mathrm{trk}}$ denote the tracking encoder and
$z_t=(z_{t,1},\ldots,z_{t,M})$ the motor-token sequence
at control step $t$.
The prior
$p_{\theta,\phi_k}(\cdot\mid z_{t,<j},o_t)$
predicts token $j$ from the preceding tokens and complete
controller input $o_t$, using task-adapter parameters
$\phi_k$.
The decoder converts the tokens and body state into
joint-position targets:
\begin{equation}
    \begin{aligned}
        u_t &= D_\psi(s_t,z_t),\\
        a_t &= \mathcal T(u_t)
             = q_{\mathrm{off}}
             + q_{\mathrm{scale}}\odot\tanh(u_t).
    \end{aligned}
    \label{eq:physical_decoding}
\end{equation}
Here $u_t$ is the dimensionless decoder output, and
$q_{\mathrm{off}}$ and $q_{\mathrm{scale}}$ are fixed
joint offsets and scales. We denote the token sampler and state-conditioned action
decoder jointly by $C_k$, as shown in
Fig.~\ref{fig:method_overview}.
Autoregression generates the $M$ motor tokens within
each control step; the token prefix is not a history
of previous control steps.
The tracking encoder $E_{\mathrm{trk}}$, together with FSQ, provides token targets for autoregressive-prior training and motion--scene adaptation, but is not required for task execution.
The motion vocabulary, original prior parameters $\theta$,
and decoder parameters $\psi$ remain fixed during
downstream adaptation. Within the controller, only the task adapters and their conditioning projections are updated.

\noindent {\bf Interaction and affordance pre-training.}

Using the G1-retargeted BONES-SEED~\cite{bones_seed} collection described in
Sec.~\ref{sec:experimental_setup}, we pair each motion with
a support scene. Motion metadata and hand/root kinematics
guide fixed-support placement around interaction intervals.
Walking clips with nearby supports or invalid geometry provide
non-interaction examples.

The interaction context $e_t$ combines task-specific inputs, support geometry and contact information, and stability features.
The support descriptor $g_t$ combines surface geometry,
contact magnitudes, and a geometry-validity indicator.
For a selected body $i$, its geometric feature is
\begin{equation}
    d_{t,i}
    =
    R_t^\top
    \big(
        p^{\mathrm{obj}}_{t,i}
        -
        p^{\mathrm{body}}_{t,i}
    \big),
    \label{eq:object_relative_geometry}
\end{equation}
where the two positions are expressed in world coordinates,
and $R_t$ maps pelvis-heading coordinates to the world frame.
During affordance pre-training rollouts, $p^{\mathrm{obj}}_{t,i}$ is the
closest sampled point on a valid support surface.
The left/right hands, wrists, and elbows provide six
displacement vectors. These are combined with six
normalized body-contact force magnitudes and one validity
flag to form $g_t$.
Displacements are scaled and clipped before conditioning
the controller.
The stability descriptor $b_t$ complements these features
with torso orientation, ground-relative height, and
linear and angular motion.

Training examples are collected through frozen-tracker
rollouts in the paired scenes.
Each input combines the current simulated body state,
support and stability descriptors, and a reference-derived
future-root planar target.
The tracker encoder and FSQ provide aligned token targets
$z_t^{\mathrm{trk}}$, which also drive the decoder during
the rollout.
Contact features are measured in simulation and may include
ground contact; they are not annotated intended-contact
labels.
This pairing associates motion-derived motor decisions
with the geometry and physical state in which they are
executed.

Following GPC~\cite{shi2026gpc}, we optimize only the
DoRA/FiLM adapters  and conditioning projections with
teacher-forced cross-entropy averaged over $M=8$ tokens
and label smoothing 0.01. Geometry, contact, and stability
serve as conditioning features, not separate SFT objectives.

\noindent {\bf Privileged task learning.}

Task reinforcement learning specializes $\phi_k$ for
Eq.~\ref{eq:interaction_objective} using GPC's token-space
optimization. Surface associations, contact definitions, and
task conditions are specialized within the same input layout.
The retained motor foundation then executes from current
state and interaction context without prescribed motion
trajectories.

\subsection{Student Distillation and Sensor Transfer}
\label{sec:sensor_transfer}

The student retains the expert's motor prior and decoder
and learns a sensor-based input interface.
Transfer proceeds in three steps.
First, observation reconstruction, motor-token matching,
and decoded-action matching train only the estimator
$F_\vartheta$ through the frozen task-adapted controller.
Second, with the estimator fixed, a bounded residual
$\Delta_\eta$ calibrates the inputs used for token selection
without modifying the body-state estimate used for decoding.
Third, task reinforcement learning updates the student
adapter and a training-only critic while keeping the
estimator, calibration map, motor backbone, decoder, and
sampling reference fixed.

\subsubsection{Structured Observation Reconstruction}
\label{sec:observation_estimation}

The structured estimator $F_\vartheta$ combines learned
perception with fixed kinematic reconstruction:
\begin{equation}
    \hat o_t = F_\vartheta(\mathcal I_t),
    \label{eq:structured_estimator}
\end{equation}
Here $\mathcal I_t$ contains the sensor history and supplied
task inputs defined in Sec.~\ref{sec:problem_formulation}.
The reconstructed controller input is
$\hat o_t=(\hat s_t,\hat e_t)$.
The estimator reconstructs body state, support features,
and stability information from onboard sensing, while
supplied task inputs are passed through unchanged.

Depth and LiDAR encoders extract local geometry, while
a recurrent proprioceptive encoder summarizes motion,
effort, tracking error, and previous actions.
Their fused features, together with task conditions,
sensor status, and ground features, predict body-relative
surface displacements, contact features, and geometry
validity.
Simulator geometry and contact measurements supervise
these predictions during training.

We reconstruct the articulated body state using known
kinematics rather than predicting every body quantity
independently.
A separate recurrent estimator predicts torso height and
heading-local linear velocity,
$\hat\xi_t=(\hat h_t,\hat{\boldsymbol v}_t)$,
from proprioceptive history and ground features.
Ground features combine LiDAR plane fitting and
forward-kinematic foot geometry.
The body and stability estimates are then
\begin{equation}
    (\hat s_t,\hat b_t)
    =
    \mathcal K(q_t,\dot q_t,Q_t,\omega_t,\hat\xi_t),
    \label{eq:analytic_lifting}
\end{equation}
where $q_t,\dot q_t$ are measured joint positions and
velocities, $Q_t$ is the pelvis orientation estimate,
and $\omega_t$ is pelvis-frame angular velocity.
The reconstruction map $\mathcal K$ accounts for the
controller's height convention, joint-induced motion,
and link-origin-to-center-of-mass velocity offsets.
Privileged body states provide training supervision but
are not inputs to the deployed body policy.

\subsubsection{Distillation Through the Retained Prior}
\label{sec:policy_distillation}

Observation reconstruction alone does not distinguish
errors that change motor decisions from those that change
physical action decoding.
We therefore supervise the estimator at all three levels:
\begin{equation}
    \mathcal L_{\mathrm{distill}}
    =
    \lambda_o\mathcal L_{\mathrm{obs}}
    +
    \lambda_z\mathcal L_{\mathrm{tok}}
    +
    \lambda_a\mathcal L_{\mathrm{act}}.
    \label{eq:distillation_objective}
\end{equation}
Here $\mathcal L_{\mathrm{obs}}$ is a weighted
Smooth-$L_1$ loss on body state, stability, interaction
features, and the base-state estimate $\hat\xi_t$.
Only estimator parameters $\vartheta$ are updated;
the task-adapted prior and decoder remain fixed.

\noindent\textit{Motor-token supervision.}
Let $z_t^E$ and $z_t^S$ be sequences generated from
privileged and estimated inputs at the same simulator state.
During distillation, the teacher and student distributions
are evaluations of the same fixed task-adapted prior:
\begin{equation}
    \begin{aligned}
        p_E(\cdot\mid z_{t,<j})
        &:=
        p_{\theta,\phi_k}
        (\cdot\mid z_{t,<j},o_t^E),\\
        p_S(\cdot\mid z_{t,<j})
        &:=
        p_{\theta,\phi_k}
        (\cdot\mid z_{t,<j},\hat o_t).
    \end{aligned}
    \label{eq:distillation_distributions}
\end{equation}
For prefix source $r\in\{E,S\}$, we align their next-token
distributions after the same prefix:
\begin{equation}
    \mathcal K_r
    =
    \mathbb E\!\left[
        \frac{\tau^2}{M}
        \sum_{j=1}^{M}
        D_{\mathrm{KL}}\!\left(
            p_E^\tau(\cdot\mid z_{t,<j}^r)
            \,\|\,
            p_S^\tau(\cdot\mid z_{t,<j}^r)
        \right)
    \right].
    \label{eq:prefix_alignment}
\end{equation}
Superscript $\tau$ denotes temperature-scaled distributions
obtained by dividing logits by $\tau>0$ before
softmax~\cite{hinton2015distilling}.
The token loss also includes expert-sequence cross-entropy:
\begin{equation}
    \begin{aligned}
        \mathcal L_{\mathrm{CE}}
        &=
        -\mathbb E\!\left[
            \frac{1}{M}\sum_{j=1}^{M}
            \log p_S(z_{t,j}^E\mid z_{t,<j}^E)
        \right],\\
        \mathcal L_{\mathrm{tok}}
        &=
        w_E\mathcal K_E
        +w_S\mathcal K_S
        +w_{\mathrm{CE}}\mathcal L_{\mathrm{CE}}.
    \end{aligned}
    \label{eq:token_loss}
\end{equation}
Teacher-prefix supervision aligns decisions along expert
token sequences. Student-prefix supervision queries both
controllers after the same prefix generated from estimated
inputs, exposing the estimator to its own token-selection
errors. Both prefix types can be evaluated at a
teacher-visited physical state; student-prefix supervision
does not imply that the student controls the simulator.
This follows the learner-sequence supervision principle
of on-policy distillation~\cite{agarwal2024gkd}, applied
within one control step.

\noindent\textit{Decoded-action supervision.}
Even matching tokens can produce different actions when
the decoder receives an inaccurate body state.
Define $\hat u_t^r=D_\psi(\hat s_t,z_t^r)$ and
$u_t^E=D_\psi(s_t,z_t^E)$.
We supervise decoder outputs and physical joint targets:
\begin{equation}
    \begin{aligned}
        \mathcal L_{\mathrm{act}}
        =\mathbb E\big[&
        \alpha_E H(\hat u_t^E-u_t^E)\\
        &+\alpha_S H(\hat u_t^S-u_t^E)\\
        &+\alpha_q H_W\big(
            \mathcal T(\hat u_t^S)-\mathcal T(u_t^E)
        \big)\big].
    \end{aligned}
    \label{eq:action_anchor}
\end{equation}
Here $H$ is a Huber loss and $H_W$ includes joint and
sample weights.
The teacher-token term isolates decoding error under
$\hat s_t$; the student-token terms supervise the actions
produced by the student's own decisions.

Teacher targets and sampled tokens are treated as constants.
Token losses backpropagate through the frozen prior into
the estimated conditioning inputs, while decoded-action
losses backpropagate through the frozen decoder into
$\hat s_t$.
Consequently, the action losses directly improve
decoder-state reconstruction; they do not differentiate
through categorical token choices.
Frozen network weights therefore do not imply that
gradients through their inputs are disabled.

Initial distillation uses teacher-controlled rollouts, with
both token-prefix types generated at the same physical states.
Calibration and refinement introduce student-controlled
episodes and query privileged supervision at visited states
in a DAgger-style process~\cite{ross2011dagger}.

\subsubsection{Prior-Input Calibration}
\label{sec:calibration_refinement}

To correct residual token-conditioning errors, we freeze the
estimator and apply a bounded residual only to the prior's
inputs, preserving the decoder's physical-state estimate:

\begin{equation}
    \begin{aligned}
        \Delta_\eta(\hat o_t)
        &=
        \epsilon_o\odot
        \tanh\!\left(
            C_\eta(\operatorname{sg}[\hat o_t])
        \right),\\
        \tilde o_t
        &= \hat o_t+\Delta_\eta(\hat o_t).
    \end{aligned}
    \label{eq:prior_calibration}
\end{equation}
The residual network $C_\eta$ is initialized to output zero,
$\operatorname{sg}$ denotes stop-gradient, and $\epsilon_o$
sets componentwise bounds.
Supplied task-conditioning fields have zero correction
bounds.
Token selection uses $\tilde o_t$, whereas the decoder
continues to receive $\hat s_t$.

GPC's reference-guided nucleus sampling~\cite{shi2026gpc}
restricts the active prior to tokens allowed by a frozen
reference. Because conditioning errors can alter this set,
calibration aligns the reference under privileged and
corrected inputs.

For prefix source $r\in\{E,S\}$, let
$p_{\mathrm{ref}}^{E,r}$ and
$p_{\mathrm{ref}}^{C,r}$ denote the reference distributions
conditioned on $o_t^E$ and $\tilde o_t$, respectively,
after prefix $z_{t,<j}^r$.
The reference includes a fixed adapted-controller snapshot
and its observation normalization.
Let $\mathcal N_{t,j}^{E,r}$ be the top-$p$ nucleus of
the privileged reference~\cite{holtzman2020nucleus}.
We encourage the corrected reference to retain probability
mass on this set:
\begin{equation}
    \mathcal L_{\mathrm{cov}}^r
    =
    -\mathbb E\!\left[
        \frac{1}{M}\sum_{j=1}^{M}
        \log
        \sum_{v\in\mathcal N_{t,j}^{E,r}}
        p_{\mathrm{ref}}^{C,r}(v)
    \right].
    \label{eq:calibration_coverage}
\end{equation}
Define $\mathcal K_{\mathrm{ref}}^r$ as in
Eq.~\ref{eq:prefix_alignment}, replacing the teacher and
student distributions with the privileged and corrected
reference distributions.
Let $\mathcal L_{\mathrm{NLL}}^r$ be the mean negative
log-likelihood of the expert token $z_{t,j}^E$ under the
active prior conditioned on $(\tilde o_t,z_{t,<j}^r)$.
The calibration objective is
\begin{equation}
    \begin{aligned}
        \mathcal L_{\mathrm{cal}}
        =
        \sum_{r\in\{E,S\}}\big(
        &\mu_r\mathcal K_{\mathrm{ref}}^r
        +\nu_r\mathcal L_{\mathrm{cov}}^r\\
        &+\zeta_r\mathcal L_{\mathrm{NLL}}^r
        \big),
    \end{aligned}
    \label{eq:calibration_objective}
\end{equation}
with nonnegative weights.
Only $\eta$ is updated; the estimator, active prior,
reference, and decoder remain fixed.

For execution, write
$p_{t,j}^S(v):=
p_{\theta,\phi_k}(v\mid z_{t,<j}^S,\tilde o_t)$.
The executed distribution $\bar p_{t,j}^S$ restricts
these probabilities to the reference nucleus computed
from $\tilde o_t$ and renormalizes them.
When their probability mass on that set is below
$10^{-3}$, our implementation uses reference probabilities
renormalized over the same set.
Privileged reference evaluations provide training targets
only; deployed sampling uses corrected student inputs.

\subsubsection{Task-Adapter Refinement}
\label{sec:task_adapter_refinement}

To optimize task return after calibration, we refine $\phi_k$
on student-controlled rollouts using GPC's token-space
PPO~\cite{shi2026gpc,schulman2017ppo}, regularized toward
the frozen privileged reference:
\begin{equation}
    \begin{aligned}
        \mathcal L_{\mathrm{RL}}
        ={}&\mathcal L_{\mathrm{PPO}}^S\\
        &+\beta\,\mathbb E\!\left[
            \frac{1}{M}\sum_{j=1}^{M}
            D_{\mathrm{KL}}\!\left(
                p_{\mathrm{ref}}^{E,S}
                \,\|\,
                p_{t,j}^S
            \right)
        \right].
    \end{aligned}
    \label{eq:student_refinement}
\end{equation}
Here $\mathcal L_{\mathrm{PPO}}^S$ is the negative clipped
PPO surrogate on student-controlled transitions.
The reference uses privileged input $o_t^E$, while
the active prior uses corrected input $\tilde o_t$;
both are evaluated after the same student prefix.
PPO likelihoods use $\bar p_{t,j}^S$, including the
low-overlap fallback, whereas the KL term compares
unrestricted categorical distributions.

Policy and privileged-critic updates use student-controlled
transitions; reference supervision may also use teacher
transitions. All components except the task adapter and
critic remain fixed. Observation and action reconstruction
losses are discontinued; the critic and privileged observations
are used only during training.

\section{Experiments}
\label{sec:experiments}

\subsection{Experimental Setup}
\label{sec:experimental_setup}

\noindent {\bf Implementation details.} We use a Unitree G1 humanoid with a 29-DoF learned body controller operating at 50 Hz with joint-level PD control. Simulation is performed in Isaac Lab at 200 Hz with a control decimation of four. For rail-grasping experiments, Dex3 hands provide an
additional 14 actuated finger joints controlled separately
from the learned whole-body policy.
In simulation, finger closure is triggered when the
selected palm establishes rail contact above 1\,N,
excluding contact with the rail underside.
The finger controller uses simulator-provided palm and
finger contact forces, together with hand--rail relative
motion, to regulate capture and holding.
Excessive contact force causes incremental opening,
and episode reset clears the grasp state.

Sensor-based students use \(96\times64\) depth images,
\(16\times180\) LiDAR rasters, 25 frames of proprioceptive
history, and onboard state estimates to reconstruct the expert
inputs defined in Sec.~\ref{sec:problem_formulation}.

\noindent {\bf Pre-Training.} Stage-3 motion--scene adaptation uses 1,024 clips,
totaling approximately 2.24 hours, selected from the
Unitree G1-retargeted motions in BONES-SEED~\cite{bones_seed}.
It contains 512 interaction clips spanning wall support,
pushing, vaulting, and ground rising, together with
512 walking clips. Interaction pre-training uses this
selected motion collection to adapt the motor prior
through frozen-tracker token supervision.

Affordance pre-training pairs each clip with a support scene
as described in Sec.~\ref{sec:object_conditioned_control}.
We use a temporal fallback when annotations are unavailable.
Walking examples include 256 scenes with nearby supports
and 256 with invalid padding geometry.

\noindent {\bf Training.} 

The iGPC students use observation, motor-token,
and decoded-action supervision, followed by prior-input
calibration and task-adapter refinement
(Sec.~\ref{sec:sensor_transfer}).
All reported student evaluations use autonomous student execution.

\noindent {\bf Evaluation protocol.} Reaching and student
ablations use 2,048 matched trials per policy, with four
evaluation seeds and 512 trials per seed. Each trial lasts
512 control steps (10.24\,s). Seeds vary evaluation conditions,
not independently trained policies. Policies share initial
robot states, rail configurations, and target-hand commands;
rail orientation, height, and robot--rail displacement vary
across trials. Training sensor corruption, reset noise, action
delay, dynamics randomization, and task-specific training
terminations are disabled, and no external pushes are applied.
Reactive stabilization and box pushing use the protocols below.

\begin{figure}[!t]
    \centering
    \includegraphics[width=\columnwidth]{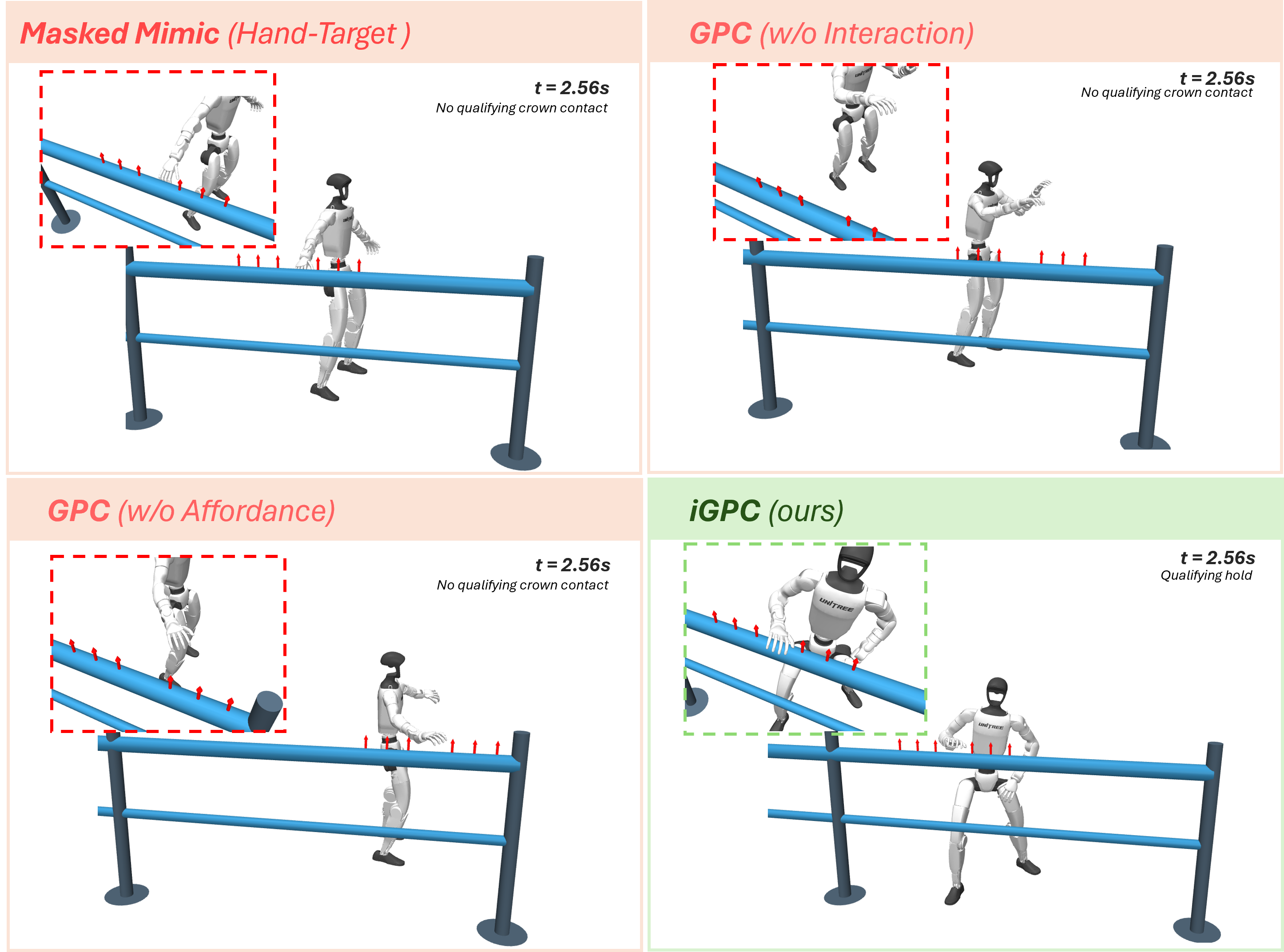}
    \caption{
        \textbf{Qualitative comparison of commanded rail-contact
        acquisition.}
        Snapshots at $t=2.56\,\mathrm{s}$ compare the policies
        evaluated in Table~\ref{reach_expert_eval}:
        MaskedMimic hand targeted to rail (top left),
        GPC (top right),
        GPC w/o support observation (bottom left),
        iGPC (ours, bottom right).
    }
    \label{fig:reaching_comparison}
\end{figure}

\subsection{Object-Aware Conditioning for Commanded Reaching}
\label{sec:reaching_experiments}

We evaluate object-aware conditioning by commanding a specified
hand to contact the upper crown of a fixed rail across varying
positions, orientations, and heights. We compare iGPC with
target-conditioned GPC, GPC adapted to interaction motions
without affordance observations, and a frozen MaskedMimic~\cite{tessler2024maskedmimic}
controller supplied with a privileged, geometry-derived
hand-target trajectory and no task-specific PPO adaptation.

To compare performance we evaluate \emph{upper-crown contact rate} that measures trials with a selected-hand crown contact of at least 15\,N. \emph{Body collision} measures pelvis, torso, or head rail contacts above 5\,N. \emph{Endpoint error} is the terminal hand-to-target distance, reported as median/p90 over all trials. \emph{Time to crown} is the time to first qualifying contact, reported over successful trials.

Table~\ref{reach_expert_eval} shows 99.32\% rail-contact
success for iGPC, compared with 31.30\% without interaction
pre-training and 48.63\% without affordance conditioning.
iGPC reduces median endpoint error from 16.03 to 1.35 cm
relative to target-conditioned GPC and reaches the rail faster.
MaskedMimic succeeds in 0.20\% of trials, so its contact-time
statistics summarize only four successes. These comparisons
evaluate complete pipelines rather than isolate pre-training
effects. Fig.~\ref{fig:reaching_comparison} illustrates the
resulting hand--rail configurations.

\begin{table}[t]
\centering
\caption{Expert policy evaluation for commanded rail-contact task.}
\label{reach_expert_eval}
\setlength{\tabcolsep}{3pt}
\renewcommand{\arraystretch}{1.15}

\resizebox{\columnwidth}{!}{%
\begin{tabular}{@{}lccrrr@{}}
\toprule
Policy &
\shortstack{Interaction\\pre-training} &
\shortstack{Affordance\\pre-training} &
\shortstack{Crown contact\\(\%) $\uparrow$} &
\shortstack{Endpoint error\\p50/p90 (cm) $\downarrow$} &
\shortstack{Time to crown\\p50/p90 (s) $\downarrow$} \\
\midrule

MaskedMimic (Hand Target)
& --- & ---
& 0.20 & 12.07/15.55 & 1.45/2.75 \\

GPC (w/o Interaction)
& $\times$ & $\times$
& 31.30 & 16.03/36.43 & 4.90/9.00 \\

GPC (w/o Affordance)
& $\checkmark$ & $\times$
& 48.63 & 15.03/32.88 & 3.93/8.44 \\

\rowcolor{gray!15}
iGPC (ours)
& $\checkmark$ & $\checkmark$
& \textbf{99.32}
& \textbf{1.35/11.04}
& \textbf{0.76/1.46} \\

\bottomrule
\end{tabular}%
}
\end{table}

\subsection{Perceptive Student Policy}
\label{sec:distillation_experiments}

We evaluate transfer to onboard sensing while retaining the pretrained GPC motor prior. We compare the privileged expert, a direct-action MLP student, and iGPC variants that ablate observation supervision, token supervision, or prior-input calibration. The MLP and no-calibration students use the same sensor
history as full iGPC and also undergo task-level PPO
refinement. The no-calibration variant retains full-loss distillation but omits the learned prior-input residual. All students execute autonomously from sensor observations.

Here, teacher-target MAE measures the mean absolute difference
between student and frozen-expert joint-position targets
after action processing, in radians. We average over
29 joints and 64 uniformly spaced samples per trial.
The same frozen expert labels states visited by each
student; this is an on-policy imitation diagnostic,
not an action comparison on identical state sequences.
As shown in Table~\ref{reach_distillation_eval}, the full iGPC student achieves 95.95\% contact success, retaining most of the expert's 99.32\% performance while outperforming the expert in terms of body collisions. In contrast, distillation of an MLP student that preserves no motion prior achieves only 46.53\% success. When ablating the design choices of our iGPC student, we observe that removing both observation and token supervision reduces success to 25.39\%, while removing either component individually yields 90.14–93.41\%. Prior calibration further improves success from 91.02\% to 95.95\% and reduces the 90th-percentile endpoint error from 13.53 cm to 5.62 cm.

\begin{table}[t]
\centering
\caption{Perception-based student policy evaluation for commanded rail-contact task.}
\label{reach_distillation_eval}
\renewcommand{\arraystretch}{1.15}
\setlength{\tabcolsep}{2.5pt}
\resizebox{\columnwidth}{!}{%
\begin{tabular}{@{}lccrrrr@{}}
\toprule
Policy &
\shortstack{Obs.\\loss} &
\shortstack{Token\\loss} &
\shortstack{Crown contact\\(\%) $\uparrow$} &
\shortstack{Body collision\\(\%) $\downarrow$} &
\shortstack{Endpoint error\\p50/p90 (cm) $\downarrow$} &
\shortstack{Teacher-target\\MAE (rad) $\downarrow$} \\
\midrule

\rowcolor{gray!15}
Expert iGPC
& --- & ---
& 99.32 & 15.19 & 1.35/11.04 & --- \\

Student MLP
& --- & ---
& 46.53 & 50.73 & 11.19/86.78 & 0.178396 \\

iGPC (no obs. or token)
& $\times$ & $\times$
& 25.39 & 17.19 & 51.78/146.20 & 0.127518 \\

iGPC (no obs.)
& $\times$ & $\checkmark$
& 93.41 & 13.13 & 1.37/8.38 & 0.088503 \\

iGPC (no token)
& $\checkmark$ & $\times$
& 90.14 & 10.55 & 2.68/16.22 & 0.088679 \\

iGPC (no prior calibration)
& $\checkmark$ & $\checkmark$
& 91.02 & 10.79 & 1.62/13.53 & 0.085493 \\

iGPC (full)
& $\checkmark$ & $\checkmark$
& \textbf{95.95} & \textbf{5.22}
& \textbf{1.22/5.62} & \textbf{0.085331} \\

\bottomrule
\end{tabular}%
}
\end{table}

\subsection{Whole-Body Interaction Experiments}

\noindent\textbf{Reactive stabilization.}
The humanoid receives two external pushes while a rail
is available for hand support. The policy must reactively
reach and grasp the rail to maintain balance, without
a prescribed reaching motion. We compare the privileged
iGPC expert, sensor-based iGPC student, and leg-only GPC
baseline using 512 matched trials in each push-induced
velocity-change band: 1.5--2.0, 2.0--2.5, and
2.5--3.0\,m/s.
A physical fall is defined as knee, pelvis, torso, or head
contact with the terrain above 10\,N for three consecutive
control frames.

\begin{figure}[!t]
    \centering
    \includegraphics[width=\columnwidth]{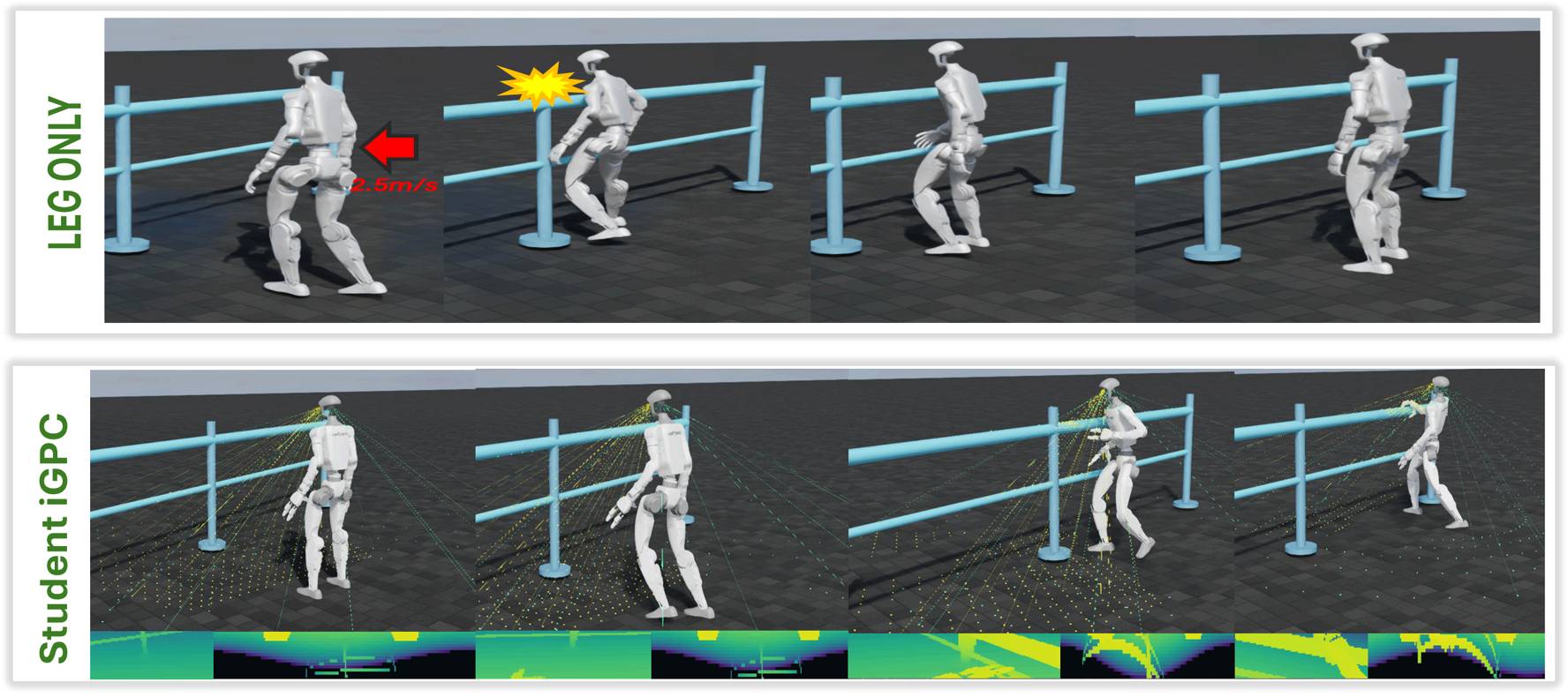}
    \caption{
        Reactive stabilization in simulation.
        Top: leg-only GPC.
        Bottom: iGPC student with sensor visualizations.
        The red arrow indicates the applied disturbance.
    }
    \label{fig:stabilization_comparison}
\end{figure}

Table~\ref{tab:whole_body_interaction}(a) reports physical
falls as counts and percentages. At 2.0--2.5\,m/s,
leg-only GPC falls in 18.36\% of trials, compared with
4.69\% for the expert and 4.88\% for the student.
At 2.5--3.0\,m/s, the student's fall rate is 17.19\%,
compared with 40.43\% for leg-only GPC.
These results support the benefit of environmental hand
support for reactive stabilization.
Fig.~\ref{fig:stabilization_comparison} illustrates the
responses of leg-only GPC and the iGPC student to
a disturbance near the rail.

\label{sec:stabilization_experiments}

\begin{table}[t]
\centering
\caption{Whole-body interaction.
(a) Physical falls, count (\%), on 512 trials per push band.
(b) Forward displacement and absolute lateral drift at first fall
or 60\,s: mean $\pm$ SD across four evaluation-seed means
(128 trials each).}
\label{tab:whole_body_interaction}

\footnotesize
\setlength{\tabcolsep}{2.5pt}
\renewcommand{\arraystretch}{1.0}
\setlength{\aboverulesep}{2pt}
\setlength{\belowrulesep}{2pt}

\begin{tabular*}{\columnwidth}{@{\extracolsep{\fill}}lrrr@{}}
\toprule
\multicolumn{4}{@{}l}{
    \textbf{(a) Reactive stabilization}: physical falls $\downarrow$
} \\
\addlinespace[1pt]
Push $\Delta v$ (m/s)
& iGPC expert
& iGPC student
& GPC leg-only \\
\midrule
1.5--2.0
& \textbf{3 (0.59\%)}
& 5 (0.98\%)
& 12 (2.34\%) \\
2.0--2.5
& \textbf{24 (4.69\%)}
& 25 (4.88\%)
& 94 (18.36\%) \\
2.5--3.0
& 116 (22.66\%)
& \textbf{88 (17.19\%)}
& 207 (40.43\%) \\

\midrule
\multicolumn{4}{@{}l}{\textbf{(b) Box pushing}} \\
\addlinespace[1pt]
Metric
& iGPC expert
& iGPC student
& VisualMimic \\
\midrule
Forward (m) $\uparrow$
& $66.08 \pm 0.15$
& $\mathbf{67.59 \pm 1.05}$
& $45.17 \pm 0.07$ \\
Lateral drift (m) $\downarrow$
& $\mathbf{12.76 \pm 1.35}$
& $13.77 \pm 1.08$
& $36.29 \pm 0.02$ \\
\bottomrule
\end{tabular*}
\end{table}
\noindent\textbf{Box pushing.}
The humanoid continuously pushes a 3.8\,kg cube over
60\,s with effective box--floor friction $\mu=0.6$.
Each policy is evaluated on four seeds with 128 trials
each, giving 512 trials per policy.
The iGPC expert runs in Isaac Lab; the iGPC student
and released VisualMimic~\cite{yin2025visualmimic}
controller run in MuJoCo.
Shared reset components and box conditions are matched,
while each method retains its native robot and control
interface.
Forward displacement is the box's net displacement
along the initial intended direction; absolute lateral
drift is its perpendicular endpoint displacement.
Both metrics are measured at the first robot fall
or 60\,s.

Table~\ref{tab:whole_body_interaction}(b) reports the mean
and standard deviation across the four evaluation-seed
means. The expert and student achieve 66.08\,m and
67.59\,m mean forward displacement, respectively,
compared with 45.17\,m for VisualMimic.
Their mean absolute lateral drift is also lower:
12.76\,m and 13.77\,m, respectively, compared with
36.29\,m for VisualMimic.
Under these evaluation conditions, the iGPC policies
achieve greater forward progress and lower lateral
drift, extending contact acquisition to sustained
whole-body object interaction.

\subsection{Deployment to the Real Robot}
\label{sec:real_robot}

The robot Unitree G1 receives depth images from an Intel RealSense
D435i camera at 30\,Hz and LiDAR measurements from a
Livox MID-360 at 10\,Hz.
The sensor-reading loop operates at 50\,Hz.
Proprioceptive inputs include joint positions and velocities,
estimated motor torques, IMU-derived orientation and angular
velocity, previous joint-position targets, joint-tracking
errors, and their history.

Policy inference runs on an NVIDIA RTX 5090 GPU, taking
approximately 18\,ms per update during hardware execution.
The controller operates continuously in closed loop,
sustaining a measured wall-clock policy-update rate above
50\,Hz. The resulting joint-position targets are sent to
the robot's low-level proportional--derivative controllers.

\begin{figure}[!t]
    \centering
    \includegraphics[width=\columnwidth]{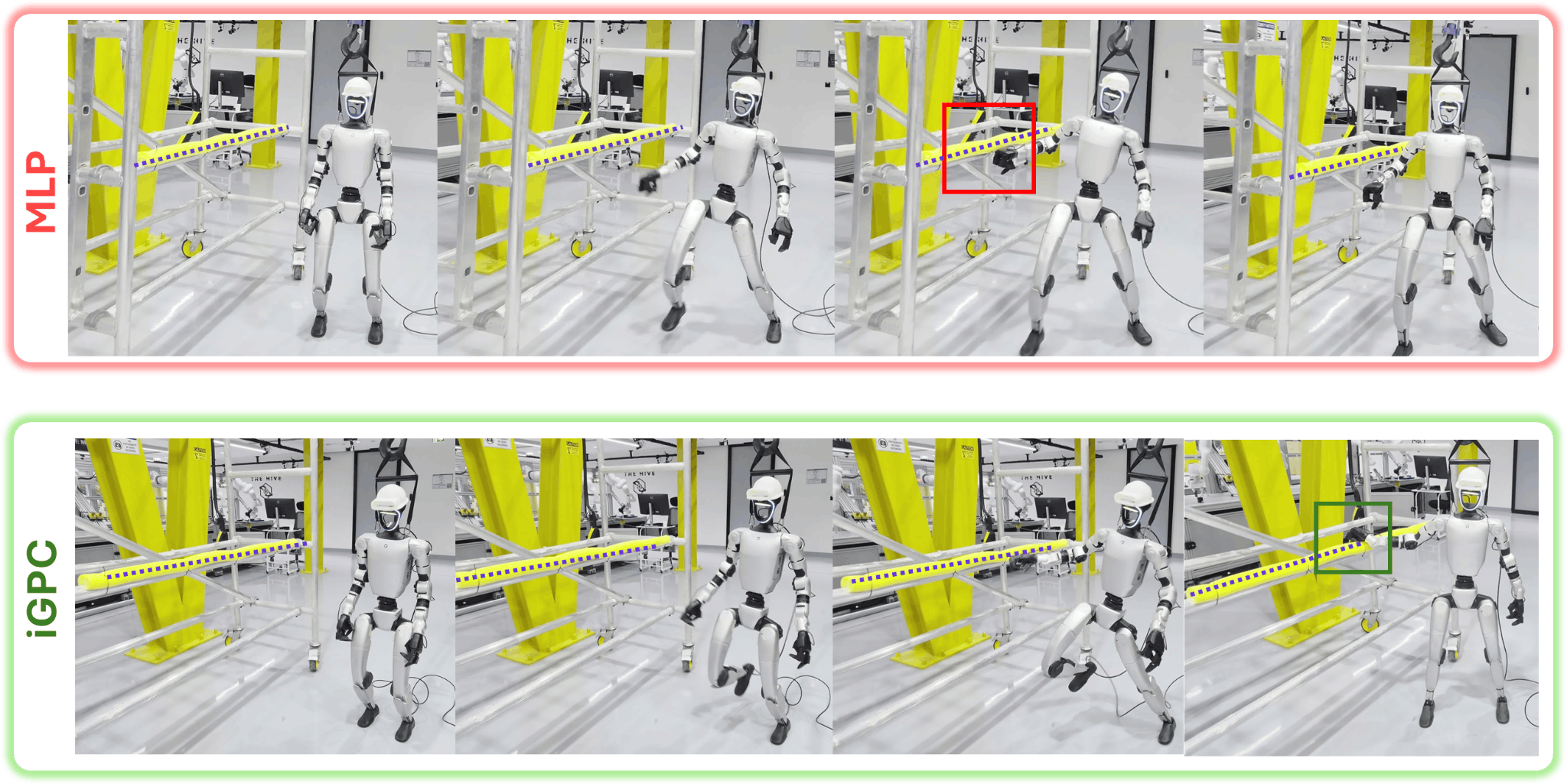}
    \caption{
        Real-robot rail-reaching comparison on Unitree G1.
        Top: direct-action MLP student.
        Bottom: iGPC student.
    }
    \label{fig:real_robot_reaching}
\end{figure}

For commanded reaching, the hand-selection condition
directs the robot to establish rail contact with the
requested hand.
Fig.~\ref{fig:real_robot_reaching} compares example
executions of the direct-action MLP and iGPC student.
In box pushing, the robot establishes contact, sustains
pushing for 10\,s, and subsequently releases contact.
For reactive stabilization, the robot grasps the rail
in response to an external push, while remaining in
stable standing when no disturbance is applied. Additionally, the robot walks to a sofa and uses hand support
to slow its descent into sitting and to rise back to standing.

\section{Conclusion}
\label{sec:conclusion}


We presented iGPC for object-aware humanoid interaction
and transfer to onboard sensing, retaining the pretrained
motor backbone and decoder. Real world deployment results support iGPC
across the evaluated reaching, stabilization, and box-pushing
tasks. Broader evaluation of generalization and contact safety
remains future work.




\section*{ACKNOWLEDGMENT}

This work was supported by the Institute of Information \&
Communications Technology Planning \& Evaluation (IITP)
grant funded by the Korean Government (MSIT)
(No.~RS-2024-00457882, National AI Research Lab Project)
and by Minerva Humanoids.

\bibliographystyle{IEEEtran} 
\bibliography{references}  

\end{document}